# Cross-view Localization and Synthesis – Dataset, Challenges and Opportunities

Ningli Xu, Rongjun Qin, *Senior Member, IEEE*

*Abstract*—Cross-view localization and synthesis are two fundamental tasks in cross-view visual understanding, which deals with cross-view datasets: overhead (satellite or aerial) and ground-level imagery. These tasks have gained increasing attention due to their broad applications in autonomous navigation, urban planning, and augmented reality. Cross-view localization aims to estimate the geographic position of ground-level images based on information provided by overhead imagery while cross-view synthesis seeks to generate ground-level images based on information from the overhead imagery. Both tasks remain challenging due to significant differences in viewing perspective, resolution, and occlusion, which are widely embedded in cross-view datasets. Recent years have witnessed rapid progress driven by the availability of large-scale datasets and novel approaches. Typically, cross-view localization is formulated as an image retrieval problem where ground-level features are matched with tiled overhead images feature, extracted by convolutional neural networks (CNNs) or vision transformers (ViTs) for cross-view feature embedding. Cross-view synthesis, on the other hand, seeks to generate ground-level views based on information from overhead imagery, generally using generative adversarial networks (GANs) or diffusion models. This paper presents a comprehensive survey of advances in cross-view localization and synthesis, reviewing widely used datasets, highlighting key challenges, and providing an organized overview of state-of-the-art techniques. Furthermore, it discusses current limitations, offers comparative analyses, and outlines promising directions for future research. We also include the project page via https://github.com/GDAOSU/Awesome-Cross-View-Methods

## I. INTRODUCTION

Cross-view imagery datasets generally refer to imagery collected from multiple sources or platforms with significantly different perspectives. Such dataset typically denotes two primary sources: overhead imagery,

The authors acknowledge the support provided by the Intelligence Advanced Research Projects Activity (IARPA) via Department of Interior/ Interior Business Center (DOI/IBC) contract number 140D0423C0034 and the Office of Naval Research (ONR, Award No. N00014-23-1-2670)."
*Corresponding author: Rongjun Qin.*
Ningli Xu is with the Geospatial Data Analytics Lab and the Department of Electrical and Computer Engineering, The Ohio State University, Columbus, OH, 43210, USA (e-mail: xu.3961@buckeyemail.osu.edu).
Rongjun Qin is with the Geospatial Data Analytics Lab, the Department of Electrical and Computer Engineering, the Department of Civil, Environmental and Geodetic Engineering, and the Transitional Data Analytics Institute, The Ohio State University, Columbus, OH, 43210, USA. (e-mail: qin.324@osu.edu).

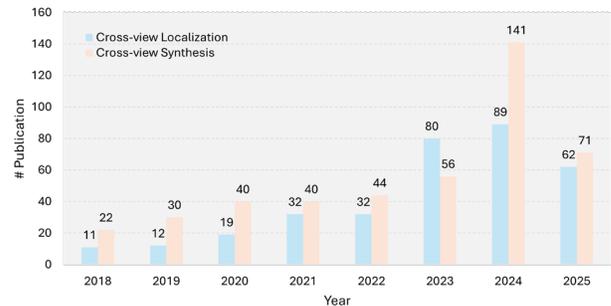

**Fig. 1.** The number of publications in cross-view localization and cross-view synthesis in past eight years.

such as satellite or aerial images, and ground-level imagery, such as data captured by vehicle-mounted or handheld cameras. These two sources form the basis of many applications in the outdoor scenes, including urban modeling [1], [2], autonomous driving [3], [4], and 3D gaming [5], [6]. Each modality offers their distinct advantages. Overhead imagery provides a broad geographic coverage of the scene content with global availability, enabling consistent observation across diverse environments. Conversely, ground-level imagery offers detailed local information about the scene, including building structures, road surfaces, and surrounding objects, which are essential for interpreting scene semantics and human-scale interactions. Despite their differences, the two sources are inherently complementary [7], [8]: overhead imagery contribute to the global spatial context, while ground-level imagery supplies fine-grained details. When integrated, these sources yield comprehensive datasets that enable accurate, high-resolution, and wide-area environmental understanding, supporting a broad spectrum of geospatial and vision-based applications.

Cross-view localization and cross-view synthesis represent two fundamental challenges in cross-view visual understanding [9], [10]. Cross-view localization aims to estimate the geospatial position of a query ground-level image by matching it to a reference database of overhead imagery. This capability is particularly critical in environments where GPS signals are unreliable or unavailable — such as urban canyons, dense forests, or indoor-to-outdoor transition zones, enabling vehicles and mobile agents to localize using purely visual cues [11]. Cross-view synthesis focuses on generating ground-level imagery conditioned on top-down view imagery, thereby bridging perspectives with substantial viewpoint differences and huge resolution differences. This task is essential for constructing immersive 3D environments, enhancing virtual and augmented reality experiences, and reconstructing content obscured or absent in overhead imagery. These two tasks advance the development of spatially aware and intelligent systems capable of interpreting and interacting with complex real-world environments across multiple viewpoints.

Existing methods for cross-view localization and synthesis often share similar underlying techniques. For example, view transformation techniques [12], [13], [14] that aim to preserve both visual appearance and spatial layout when bridging the gap between overhead and ground-level imagery. These techniques are essential for learning meaningful cross-domain correspondences despite the drastic changes in viewpoint, scale, and scene composition. However, despite their shared foundation, the network architecture used for localization and synthesis differ significantly due to the distinct objectives of these two tasks.

Cross-view localization is typically framed as an image retrieval or matching problem. As such, it relies on discriminative models, often using convolutional neural networks (CNNs) [15] or transformer-based architectures [16], [17] as backbones to extract robust features. A contrastive learning objective [18] is commonly employed to ensure that matching image pairs across views are embedded closely in feature space while non-matching pairs are pushed apart with high penalty in the loss function. This facilitates the identification of the most likely overhead reference image corresponding to a given ground-level image query. Cross-view synthesis is approached as a conditional image generation task. It leverages generative models—such as Generative Adversarial Networks (GANs) [19], [20] and diffusion models [21]—that are capable of producing realistic ground-level imagery from overhead inputs. These models focus on learning complex data distributions and producing high-fidelity outputs that capture plausible textures, structures, and semantics from a radically different viewpoint. This architectural divergence highlights the contrasting nature of the two problems: localization emphasizes discriminative feature learning and similarity measurement, while synthesis centers on generative modeling and content creation.

The structure of this paper is organized as follows. Section 2 introduces background information on both cross-view localization and synthesis. It describes key challenges in this domain (Section 2.1), reviews the relevant datasets (Section 2.2), and discusses common evaluation metrics (Section 2.3). Section 3 presents a structured overview of cross-view localization methods. We begin by outlining the general network architecture and foundational techniques, followed by a categorization of existing approaches into CNN-based methods (Section 3.1), transformer-based methods (Section 3.2), and foundation-based methods (Section 3.3). Section 4 focuses on cross-view synthesis. After introducing the fundamental concepts, we classify the methods into four categories based on their modeling paradigms: end-to-end methods (Section 4.1), voxel-based methods (Section 4.2), mesh-based methods (Section 4.3), and neural representation-based methods (Section 4.4). Section 5 discusses current limitations and open research opportunities in both cross-view localization and synthesis. Finally, Section 6 concludes the paper with a summary of key insights and future directions. We hope that this survey serves as a useful reference for researchers and practitioners interested in this rapidly evolving field.

In this work, we present a comprehensive survey of cross-view localization and synthesis, with two key contributions:

1. In light of the recent surge in research on cross-view techniques and vision architecture, we provide an up-to-date and thorough review of cross-view localization and synthesis literature, aiming to help readers quickly grasp the foundational concepts and current landscape of this field.
2. We introduce unified frameworks for both cross-view localization and synthesis tasks to guide researchers in identifying relevant approaches and better understanding the associated methodologies.

This survey aims to systematically summarize and categorize the core methodologies and resources in the field of cross-view localization and synthesis. The majority of the referenced works are published in leading photogrammetry and computer vision conferences and journals, along with recent preprints available on arXiv, where the number of publications in related field is summarized in **Fig. 1**. Rather than detailing every sub-category, we highlight representative methods to illustrate the main paradigms. For more in-depth discussions, we refer readers to the cited works.

While visual localization, view synthesis, neural rendering are conceptually related to cross-view localization and synthesis, they fall outside the scope of this review. For comprehensive overviews of those topics, we recommend [22] for visual localization, [23], [24] for view synthesis and [25], [26] for neural rendering. Our primary focus is on approaches that transform overhead (e.g., satellite or aerial) imagery into images at the ground level, as well as the generative techniques that condition on these transformed representations. Readers interested in specific generation models can refer to [20] for GAN-based methods and [27] for diffusion models. General surveys on image generation are available in [28]. The most closely related review is [9], [11], [10], where [9], [11] focuses on specifically on geo-localization with visual data and [10] focuses specifically on satellite-to-street view synthesis. In contrast, our work provides a systematic and broad review of cross-view localization and synthesis and proposes a general framework to support future research in this area.

## II. BACKGROUND

### A. Challenges

Cross-view localization and synthesis share several fundamental challenges stemming from the inherent characteristics of cross-view datasets, while each task also faces its own unique difficulties.

**Table 1.** A summary of existing datasets for cross-view localization and cross-view synthesis. "Sparsity" means the collection distance between neighboring ground-level images. "PAN" is short for panorama format, "PERS" is short for perspective format. "GSV" is short for Google Street View. "Bing" is short for Bing Map, "Google" is short Google Map. "Source(G), Source(O)" means data source for ground-level and overhead imagery.

| Cross-view Localization | | | | | | | |
|---|---|---|---|---|---|---|---|
| **Dataset** | **Year** | **#Images** | **#Cities** | **GPS** | **Orientation** | **Source (G)** | **Source(O)** |
| CVUSA[31] | 2015 | 44,416 | 20 | Yes | - | GSV&Flicker | Bing |
| UrbanGeo[36] | 2017 | 8,851 | 3 | Yes | Heading | GSV | Google |
| CVACT[32] | 2019 | 128,334 | 1 | Yes | - | GSV | Google |
| VIGOR[37] | 2021 | 105,214 | 4 | Yes | Heading | GSV | Google |
| **Cross-view Synthesis** | | | | | | | |
| **Dataset** | **Year** | **#Images** | **#Cities** | **Sparsity** | **Format** | **Source (G)** | **Source(O)** |
| CVUSA[31] | 2015 | 44,416 | 20 | Sparse | PAN | GSV&Flicker | Bing |
| Dayton[38] | 2016 | 76,084 | 11 | Sparse | PERS | GSV | Google |
| CVACT[32] | 2019 | 128,334 | 1 | Sparse | PAN | GSV | Google |
| Sat2GS[34] | 2025 | 99,825 | 1 | 10 m | PAN&PERS | GSV | DFC2019 |

A defining feature of cross-view datasets is the drastic difference between ground-level and overhead imagery in terms of perspective, resolution, and occlusion, which introduces substantial complexity for both tasks. For instance, the overhead imagery typically captures a wide-area layout, whereas the ground-level imagery offers a human-eye-level perspective rich in scene details but limited in spatial context. This severe viewing angle difference, along with differences in image resolution and frequent occlusions from buildings, vegetation, and other structures, poses significant challenges for establishing spatial or visual correspondence. Consequently, both localization and synthesis approaches must have robust strategies to bridge these domain gaps—such as learning view-invariant representations or inferring missing information—while also addressing their task-specific demands.

- **Viewing perspective difference**. There is nearly 90 degree viewing angle difference between the overhead and ground-level imagery. The scene layout and appearance information are stored in a completely different format. It will add to the difficulty of learning the robust and general mapping relation between cross-view pairs.
- **Occlusions**. Overhead images of the scene capture the building roof and ground surface while the ground-level images mainly capture the building facades and ground surface. The overlap between cross-view pairs is around ground surface, which is featureless regions and makes it difficult for localization methods to extract the features. In addition, the lack of visual cues in the occluded region may incur higher level of hallucination when exploring conditional generation methods [29], [30].
- **Resolution difference**. There is nearly 10 times difference in the resolution between cross-view pairs, where the GSD of commercial satellite images is 0.5m/pixel and the ground-level imagery usually has the GSD of 5cm/pixel [30]. For localization tasks, the scale difference makes it difficult to extract robust features to construct the matches. For example, a car accounts for tenths of pixels in satellite images and accounts for hundreds of pixels in ground-level images. Such large-scale difference often fails both traditional and learning-based feature extractors. For synthesis tasks, synthesizing high-resolution images that is 10 times resolution than the input images are beyond the image super-resolution task, not to mention the viewing perspective difference.

In practical scenarios, several factors hinder model performance. Cross-view applications typically involve large-scale databases spanning vast geographic regions, often containing thousands or even millions of high-resolution overhead image tiles [31], [32]. Managing such massive datasets requires substantial computational power and memory resources, making it essential to balance retrieval speed and matching accuracy. Another major challenge lies in extracting meaningful visual cues from top-down images to guide the generation of high-quality ground-level views. Overhead imagery generally provides only coarse structural layouts with limited appearance details, making it difficult to infer fine-grained features such as textures, building facades, vegetation layouts, and landscape styles [30]. Many of these elements are either occluded or entirely missing, forcing the generative model to plausibly infer them in a context-aware manner. Furthermore, the characteristics of these missing details vary widely across geographic regions due to differences in architecture, vegetation, and urban design. Consequently, models must generalize across diverse environments while maintaining photorealism and structural consistency in the generated images [33], [34].

*B. Datasets*

There exists a number of datasets for cross-view tasks. **Table 1** lists some typical datasets used for cross-view localization, and cross-view synthesis. Since cross-view localization and synthesis are sibling tasks, thus, some datasets used for one could be used for the other, such as CVUSA [31], CVACT [32]. It is evident that most existing datasets share similar data sources, with ground-level images primarily obtained from the Google Street View API [35] and overhead images mostly sourced from the Google Maps or Bing Maps APIs except Sat2GroundScape [34]. This reliance stems from the challenges associated with collecting large-scale ground-level and overhead imagery, and alignment of cross-view data. We introduce the details of each dataset below.

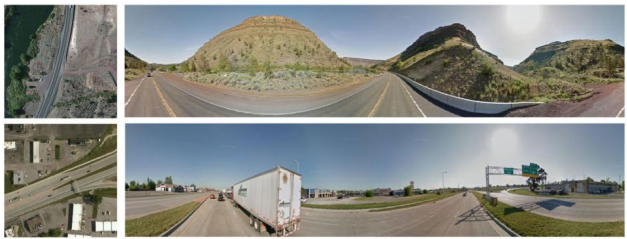

**Fig. 2.** Sample views from CVUSA dataset. The paired views are distributed across the whole United States.

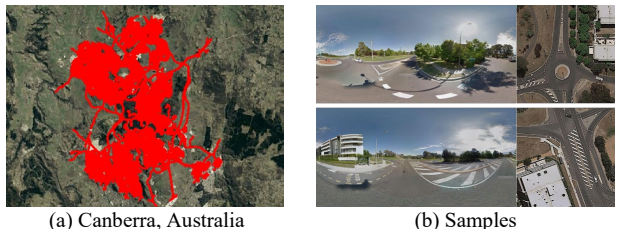

(a) Canberra, Australia          (b) Samples

**Fig. 3.** Location and samples from CVACT dataset.

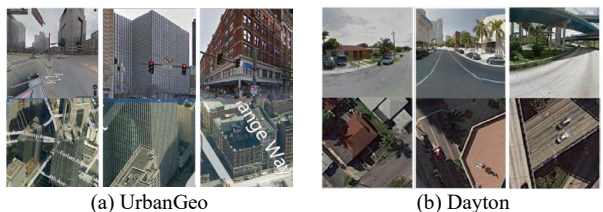

(a) UrbanGeo          (b) Dayton

**Fig. 4.** Examples of the UrbanGeo and Dayton datasets. Top row represents the ground-level images and bottom row represents the aerial views.

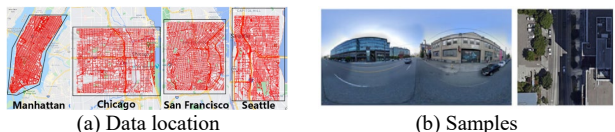

(a) Data location          (b) Samples

**Fig. 5.** Location and samples of the VIGOR dataset.

For cross-view localization datasets, they are designed to evaluate the retraval performance where each ground-level image serves as a query, and the goal is to identify the best matching overhead image from a set of candidates. These datasets typically provide multiple overhead images as potential matches, forming a database against which the ground-level queries are compared. The primary purpose is to evaluate the retrieval performance of different methods under diverse geographic and visual conditions. The broader the geographic coverage of the dataset, the more challenging the retrieval task becomes. Below, we introduce the dataset in detail.

**CVUSA** [31], short for Cross-View USA, is a commonly used dataset in the field of cross-view geo-localization and synthesis. It consists of 44,416 pairs of satellite and panoramic ground-level images covering across the U.S. The images are sourced from Flickr and Google Street View, with the street-view panoramas having a resolution of 1232 × 224 pixels, as shown in **Fig. 2**. For Flickr, the country is divided into a 100 × 100 grid, and approximately 150 images per grid cell have been collected since 2012. Google Street View images are randomly sampled from the selected locations across the U.S.

**CVACT** [32], short for City-Scale Cross-View Dataset for the Australian Capital Territory, contains 128,334 pairs of satellite and panorama ground-level images collected across Canberra City, Australia. The dataset includes 35,532 pairs for training and 92,802 for testing, as shown in **Fig. 3**. The street-view images are retrieved using the Google Street View API. Corresponding satellite images are obtained from the Google Maps API, where each image centered at the GPS location of its street-view counterpart and downloaded at a resolution of 1200 × 1200 pixels. **Fig. 3** provides sample image pairs from the dataset.

**UrbanGeo** [36] focuses on cross-view geo-localization in urban environments, with data collected from three U.S. cities: Pittsburgh, Orlando, and part of Manhattan. The author gathered 8,851 GPS locations across these cities, as shown in **Fig. 4-(a).** For each location, four bird's-eye view (BEV) images were collected at four directions 0°, 90°, 180°, and 270°. Using the DualMaps20 tool, corresponding street-view images from Google Street View were obtained based on the BEV. Furthermore, manual efforts were used to annotate building outlines in both the street-view and BEV images when the same building appeared in both viewports.

**VIGOR** [37] is a large-scale benchmark designed for cross-view geo-localization task. It includes aerial images and ground-level panoramas collected from four major U.S. cities: Chicago, New York City, Seattle, and San Francisco, as shown in **Fig. 5**. Aerial images were obtained using Google Maps Static API, while street-level panoramas were collected via Google Street View Static API. Unlike CVUSA and CVACT, which are primarily used for image retrieval tasks, VIGOR is tailored for fine-grained geo-localization. A distinguishing feature of VIGOR is its regular spacing between panorama capture locations—typically around 30 meters—ensuring consistent spatial sampling. To maintain dataset balance, each overhead image is linked to at most two panorama images. The overhead images have a resolution of 640 × 640 pixels, while the panoramas are 2048 × 1024 pixels. Satellite views are captured at zoom level 20, approximately equating to a ground resolution distance of 0.1 meters. Additionally, each panorama is oriented with North at the center to standardize directional alignment. Around 4% of the overhead images intentionally lack corresponding panorama images and serve as distractor samples, making the dataset more challenging.

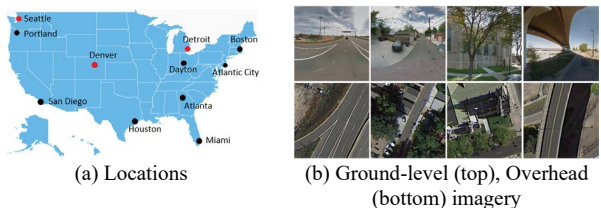

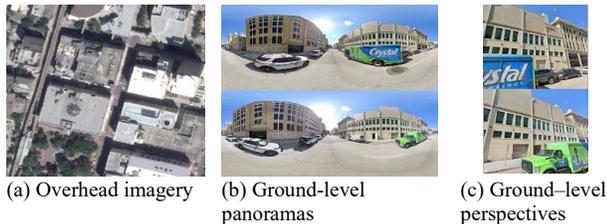

**Fig. 6.** Locations and sample views from Dayton dataset.

**Fig. 7.** Location and samples from Sat2GroundScape dataset.

For cross-view synthesis datasets, they are designed to provide one-to-one pairs of overhead and ground-level images, where the overhead image serves as input and the ground-level image serves as ground truth. These datasets are used to evaluate the quality of synthesized ground-level imagery, focusing on photorealism and structural similarity. CVUSA and CVACT, previously described, are also widely used for cross-view synthesis tasks. Below, we introduce other commonly used datasets.

**Dayton** [38] is a benchmark dataset designed for ground-to-aerial and aerial-to-ground image translation, commonly used in cross-view image synthesis tasks. It contains 76,048 pairs of satellite and ground-level images collected from 11 cities across the U.S. The street-view images are sourced from Google Maps, where multiple crops are extracted from each panoramic view. Corresponding satellite images are then retrieved via Google Maps at a resolution of $354 \times 354$ pixels. **Fig. 6** displays example image pairs from the dataset.

**Sat2GroundScape** [34] is the first satellite-to-ground dataset offering dense ground-level data suitable for ground video generation applications, as illustrated in **Fig. 7**. The dataset comprises 99,825 pairs of perspective-format images and 25,503 pairs of panoramic-format images collected in Jacksonville, Florida, USA, with spatial intervals ranging from 3 to 10 meters. The satellite imagery is sourced from the 2019 Data Fusion Contest dataset [39], which provides high-resolution, multi-view satellite images along with metadata. Ground-level images are obtained from Google Street View, with a resolution of $512 \times 1024$ pixels. Satellite and ground-level imagery are precisely aligned through manual adjustments accounting for longitude, latitude, elevation, and orientation.

### C. Metrics

Different evaluation metrics have been proposed to test these methods for various cross-view tasks.

In cross-view localization task, the objective is to assess how accurately the predicted geo-location matches the ground-truth location. In practice, since ground-truth locations are embedded in GPS-tagged overhead images, evaluation typically focuses on retrieval performance - specifically, whether the correct overhead image corresponding to a query ground-level image is ranked within the top-k predictions. More recently, methods have also begun to assess meter-level localization accuracy by measuring the distance between predicted and actual coordinates. The details of the evaluation metrics are introduced below.

**Top-k recall accuracy and hit rate** [37]. For each query image, the top-k nearest overhead images is retrieved as predictions. A retrieval is considered correct if the ground-truth image is included among the top-k results. If the top-1 retrieved reference image covers the query location (including the ground-truth), it is counted as a hit, and the hit rate is reported as part of the retrieval evaluation.

**Meter-level accuracy.** It evaluates localization accuracy by setting multiple distance thresholds (such as 0, 36, 72, and 144 meters) and measuring the error between the predicted and ground-truth GPS locations. A prediction is considered successful if the localization error falls below the specified threshold. Model accuracy is then computed based on the proportion of successful localizations. Among these thresholds, the 0-meter criterion is the most challenging, as it requires an exact match between the predicted and actual locations.

In cross-view synthesis task, the goal is to evaluate the quality of the synthesized ground-level image relative to the ground-truth image, focusing on photorealism and structural similarity. Standard low-level image similarity metrics such as PSNR and SSIM [40] can be applied for this task. Additionally, perceptual quality metrics that align with human visual judgment—such as LPIPS [41], FID [42], and DreamSIM [43]—are also employed. The following section provides detailed descriptions of these metrics.

**Peak Signal-to-Noise Ratio (PSNR)** measures the ratio between the maximum possible signal power and the noise that causes distortion, impacting the accuracy of signal representation [67]. In this context, PSNR quantifies the similarity between two images and is expressed in decibels (dB). It is defined as follows:

$$PSNR = 10 log_{10} \frac{peakval^2}{MSE} \qquad (1)$$

**Table 2.** Some representative methods in cross-view localization. We show the three common components in cross-view localization methods, including the cross-view technique, feature backbone, and loss. "CNN*" means the specially designed CNN architecture. "CE" means the cross-entropy loss. "DA" means the domain adaptation loss.

| Method | Publication | Cross-view technique | Feature Backbone | Loss |
|---|---|---|---|---|
| DSTG[44] | 2025 TGARS | - | Swin Transformer | Triplet |
| GeoDistill[45] | 2025 arXiv | Spherical Transform | DINOv2 | CE |
| CV-Cities[46] | 2025 JSTARS | - | DINOv2 | InfoNCE |
| AuxGeo[47] | 2025 ISPRSJ | Spherical Transform | ConvNeXt | InfoNCE |
| VFA[12] | 2024 CVPR | Plane Homography | U-Net | DA |
| CAMP[48] | 2024 TGARS | - | ConvNeXt | InfoNCE |
| DAC[49] | 2024 TCSVT | MLP | ConvNeXt | InfoNCE |
| UCVGL[50] | 2024 CVPR | Spherical Transform | ConvNeXt | InfoNCE |
| GeoDTR+[51] | 2024 TPAMI | - | ConvNeXt | Triplet |
| Zhang et al[52] | 2023 WACV | - | VGG | Triplet |
| Sample4Geo[53] | 2023 ICCV | - | ConvNeXt | InfoNCE |
| Shi et al[14] | 2022 CVPR | Plane Homography | CNN* | L1 |
| TransGeo[54] | 2022 CVPR | - | ViT | Triplet |
| PLCD[55] | 2022 TMM | - | ResNet | Triplet |
| SIRNet[56] | 2022 TGARS | Polar Transform | VGG | Triplet |
| Hu et al[57] | 2022 MM | Polar Transform | VGG | Triplet |
| GTFL[58] | 2021 ICCV | - | VGG | Triplet |
| L2LTR[59] | 2021 NeuIPS | Transformer | ResNet | Triplet |
| VIGOR[37] | 2021 CVPR | - | VGG | Triplet |
| LPN[60] | 2021 TCSVT | - | ResNet | CE |
| EgoTR[61] | 2021 arXiv | - | ViT | Triplet |
| DSM[13] | 2020 CVPR | Polar Transform | VGG | Triplet |
| CVFT[62] | 2020 AAAI | Optimal transport | VGG | Triplet |
| Regmi et al[63] | 2019 ICCV | GANs | CNN* | Triplet |
| SAFA[64] | 2019 NeuIPS | Polar Transform | VGG | Triplet |
| FCANet[65] | 2019 ICCV | - | ResNet | Triplet |
| CVM-Net[66] | 2018 CVPR | NetVLAD | FCN | Triplet |
| Vo et al[38] | 2016 ECCV | - | AlexNet | Triplet |

**Structural Similarity Index Method (SSIM)** measures the similarity between an original image and its reconstructed version. SSIM [67] interprets image degradation as changes in the perceived structural information. It also incorporates other important perceptual factors, such as contrast masking—where distortions are less noticeable in textured regions—and luminance masking—where distortions are less visible near image boundaries. SSIM is mathematically expressed by combining contrast $c$, luminance $l$, and structural $s$ components as follows:

$$SSIM(x,y) = [l(x,y)]^{\alpha} \cdot [l(x,y)]^{\alpha} \cdot [l(x,y)]^{\alpha} \quad (2)$$

$$l(x,y) = \frac{2\mu_x\mu_y + C_1}{\mu_x^2 + \mu_y^2 + C_1} \quad (3)$$

$$c(x,y) = \frac{2\sigma_x\sigma_y + C_2}{\sigma_x^2 + \sigma_y^2 + C_2} \quad (4)$$

$$s(x,y) = \frac{\sigma_{xy} + C_3}{\sigma_x\sigma_y + C_3} \quad (5)$$

**Frechet Inception Distance (FID)** measures the distance between the real data distribution $p_{real}(x)$ and the generated data distribution $p_{g(x)}$ based on extracted visual features [68]. FID models both distributions as multivariate Gaussians and calculates the distance between them.

**Learned Perceptual Image Patch Similarity (LPIPS)** leverages deep learning to measure perceptual differences between images. It employs a deep convolutional neural network, pretrained on image classification tasks, to extract feature representations from images. The perceptual similarity is then quantified by calculating the distance between these feature vectors. LPIPS is flexible and can utilize various neural network architectures [41], with experiments conducted using popular models such as SqueezeNet [69], AlexNet [70], and VGG [71].

**DreamSIM** [43] is a recently introduced metric designed to better align with human perception. It emphasizes comparing mid-level similarities and differences in image layout, object pose, and semantics in a holistic manner.

## III. CROSS-VIEW LOCALIZATION

The core input to cross-view localization is a ground-level view as the query, termed as $Q$. This query captures visual features of a specific location from the ground-level perspective, although its precise geographic locations are unknown. In addition, there exists a collection of overhead images $A_1, A_2, ..., A_N$, where each $A_i$ is linked to geographic location $(x_i, y_i)$. These overhead images offer a top-down perspective of different locations and serve as reference candidates for matching with the query image $Q$. The objective is to estimate the location $(x^*, y^*)$ of the query $Q$ by identifying the overhead image that most closely matches it in

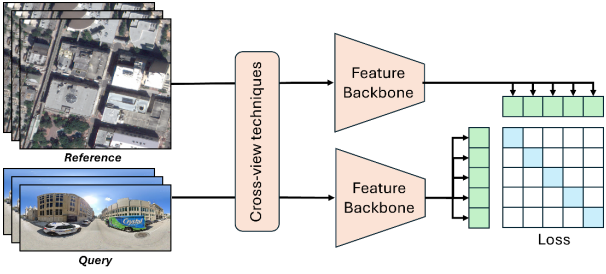

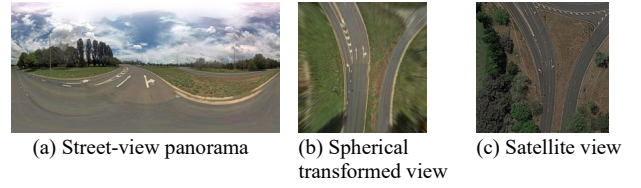

(a) Street-view panorama    (b) Spherical transformed view    (c) Satellite view

**Fig. 9.** Samples of the spherical transform. Images courtesy from [80].

**Fig. 8.** An illustration of a typical network architecture used in cross-view localization methods. It follows the standard Siamese network architecture [49], where two identical subnetworks share the same weights and extract the feature representations from two difference inputs. Additionally, various cross-view techniques are also proposed to bridge the domain gap between cross-view datasets. The goal of loss function is to drive the network to generate smaller distances for similar pairs and larger distances for dissimilar ones.

appearance. This process can be formally represented by the following equation:

$$(x^*, y^*) = arg \max_{(x_i, y_i)} S(Q, A_i) \qquad (6)$$

where $S(Q, A_i)$ represents a similarity function that quantifies the distance between the ground-level query $Q$ and the overhead image $Q$.

The similarity function $S(Q, A_i)$ has to be carefully designed or learned to robustly capture the correspondence between ground-level and overhead imagery. Ideally, it should assign higher similarity scores to image pairs that correspond to the same geographic location.

Existing methods generally follow a standard Siamese network architecture [72], [73], [74], as shown in **Fig. 8**. It first applies cross-view techniques to bridge the domain gap between cross-view dataset, then use two identical feature backbones [75], [76] to extract features and finally use various loss functions to distinguish the ground truth cross-view pair from potential candidates. This special architecture is designed to learn similarity or distance metrics between pairs of input instances [18]. Their primary goal is to analyze and discern relational attributes between different instances of data.

**Triplet loss** is commonly used as a variant of contrastive loss to enhance model performance [77], [78], [79]. Its goal is to learn distinct feature representations by evaluating the relationships among three inputs: an anchor, a positive, and a negative example. The objective is to reduce the distance between the anchor $A$ and the positive $P$ in the feature space while increasing the distance between the anchor and the negative $N$, ensuring the separation exceeds a predefined margin, as shown below:

$$L_{triplet} = \max(0, D(A,P) - D(A,N) + margin), \qquad (7)$$

where, $D(A,P)$ represents the distance between the anchor $A$ and the positive instance $P$, while $D(A,N)$ denotes the distance between the anchor and the negative instance $N$. The margin is a predefined threshold that specifies how much larger $D(A,N)$ should be compared to $D(A,P)$ for the loss to be minimized. If the difference between these two distances exceeds the margin, the loss becomes zero, indicating that the network has successfully learned to distinguish between the instances.

The common used technique to bridge the domain gap between cross-view datasets is **spherical transform** [45], [50], [80], [81]. It is a projection from ground-level panorama to top-down view image, as shown in **Fig. 9**. To obtain the corresponding overhead view of a panorama, a tangent plane is placed at the south pole of the spherical imaging surface, which serves as the new imaging plane. Let the panorama have a resolution of $H_p \times W_p$, and let the transformed output be $H_b \times W_b$ after applying the spherical transform. For each pixel on the top-down imaging plane, the camera optical center is connected to the pixel and determines its corresponding location on the panorama by finding the intersection of this ray with the spherical imaging surface. Let the top-down view pixel coordinates be denoted as $(u_b, v_b)$, and the corresponding 3D camera coordinates as $(x_1, y_1, z_1)$, which can be directly computed. The field of view, named as $fov$, for the top-down view is set to a certain value (such as $fov = 85$ degree), and the focal length used in the projection process is defined as $f = 0.5W_b/\tan(fov)$. Using this setup, we can establish the corresponding pixel coordinates $(u_p, v_p)$ on the panoramic image, as shown below:

$$u_p = 0.5[1 - arctan2(W_b/2 - u_b, H_b/2 - v_b)/\pi]W_p \qquad (8)$$
$$v_p = [0.5 - arctan2(-f, d)/\pi]H_p \qquad (9)$$

where $d = \sqrt{(W_b/2 - u_b)^2 + (H_b/2 - v_b)^2}$ represents the distance between camera center and the overhead image center.

After applying the spherical transform, the ground-level image is projected into the overhead view. This method enables high-resolution localization without the need for multiple sampling steps or dividing the satellite image into smaller patches.

We list the representative methods in **Table 2**. Based on the types of their feature backbone, we can categorize the existing cross-view localization methods into convolution-based methods, transformer-based methods, foundation-based methods. Convolution-based methods use CNN architectures (such as ResNet [75], [82], [83], VGG [71], [84], or ConvNeXt [85], [86], [87]) as backbones to extract feature representations, while transformer-based methods adopt vision transformer (ViT) architectures [76], [88], [89] for the same purpose. Convolution-based methods perform well with limited training data due to their strong inductive biases but tend to generalize poorly to unseen regions or data. In contrast, transformer-based methods excel when trained on larger datasets and offer better generalization. Foundation-based methods leverage large pre-trained vision foundation models (e.g., Swin Transformer [90], [91], [92] or DINO [93], [94], [95], [96]) as feature backbones and use distillation techniques to extract relevant knowledge for cross-view localization.

### A. Convolution-based methods

For a long time, CNNs have been the primary foundation for learning effective representations in the cross-view geo-localization research community [38], [63], [65]. These methods apply two identical convolution networks with sharing weights as the backbones to extract the local features from both ground-level and overhead images. Since convolution-based networks tend to learn the local features, how to aggregate the local features from cross-view domains into a unified feature space is a key challenge for these methods.

The use of CNN as the feature backbone in cross-view geo-localization dates back to 2016. Vo et al [38] explore several Siamese architectures with various CNN network configurations. They introduced a novel distance based logistic (DBL) layer in addition to the standard contrastive loss in order to tackle the rotation invariance problem faced by CNN and show that the proposed DBL layer has significantly improved representation learning networks.

Deep convolutional networks, such as VGG [71] and ResNet [75], are also used for this task. FCANet [65] applied ResNet-34 as the feature backbones to learn the local feature. It proposed a lightweight attention module, integrated into ResNet-34, to improve the feature discriminativeness and learn the context-aware feature. LPN [60] used ResNet-50 as the backbones. In order to learn the global context feature, it introduced a square-ring feature partition strategy that allocates attention based on the distance from the image center. Zhang et al. [52] applied VGG16 pretrained on ImageNet [97] as backbones of feature extractors. It introduced a novel temporal feature aggregation method that learns end-to-end feature representations from sequences of limited field-of-view (FOV) images for sequence-based geo-localization.

A modern convolutional network architecture is proposed using the design principles from vision transformers in 2022, called ConvNeXt [85]. The following cross-view localization methods started to switch to ConvNeXt as feature backbones [50], [51], [53], [80]. Sample4Geo [53] applies ConvNeXt-B with sharing weights as the feature backbone. It proposed two sampling strategies for improving the training process, a GPS-based sampling strategy designed to select hard negatives at the start of training, combined with a dynamic similarity sampling method that continuously identifies hard negatives throughout the training process. Similarly, AuxGeo [47] applied ConvNeXt-B as the feature backbone, and proposed the position constraint module (PCM) to enhance the feature extraction performance by using position prior information to aggregate the coarse and fine-grained features.

### B. Transformer-based methods

ViT [76], [88], [89] has achieved significant performance on various vision tasks due to its powerful global modeling ability and self-attention mechanism. The use of ViT for cross-view localization tasks can explicitly encode the position information, thus can directly learn the geometric correspondence between two views with the learnable position embedding.

TransGeo [54] is a pioneering work that applied pure transformer-based architecture for cross-view localization task. It solved the limitation of original ViT on extremely large training data size and memory consumption by incorporating adaptive sharpness aware minimization. It avoids overfitting to local minima by optimizing the adaptive sharpness loss landscape and improves model generalization performance. Dai et al. [98] designed FSRA to solve the problem of position offset and uncertainty of distance and scale. It introduced a feature segmentation and region alignment structure to enhance contextual understanding. EgoTR [61] applies ViT to extract the global features from cross-view pairs. It introduced a novel self-cross attention mechanism that enables interaction among cross-layer patches, facilitating efficient information flow across Transformer blocks and promoting continuous refinement of feature representations.

### C. Foundation-based methods

Recently, vision foundation models have shown outstanding performance across a wide range of visual tasks [99], [100], [101]. Trained on large and diverse datasets, these models are capable of capturing image details and features more effectively than traditional approaches. Their strong generalization capabilities lead to robust performance across different tasks, and they support a broader range of applications compared to earlier models. Existing methods directly borrow the off-the-shell foundation model as the feature backbone or distill the knowledge from pre-trained models for cross-view localization tasks[44], [45], [46].

CV-Cities [46] applies vision foundation model DINOv2 [94] as the feature backbone and a MLP-based mixture module to aggregate the extracted features from cross-view datasets. It also proposed two sampling strategies, near-neighbor sampling and dynamic similarity sampling, to mine negative samples for model training to improve the accuracy of the framework localization. GeoDistill [45] also applied a pretrained DINOv2-b14 as the feature extractor. Instead of borrowing the off-the-shell model, it distills the knowledge from the pre-trained model by introducing lightweight DPT module [102] and during training process, the DINOv2 weights are frozen and the DPT module are updated to fine tune the DINO feature. DSTG [44] distill the knowledge from Swin Transformer [90] to improve the cross-view localization performance. The hierarchical feature extraction of ST enables the capture of semantic information at multiple scales, enhancing feature alignment between teacher and student models at different resolution levels. It proposed a lightweight distilled model DSTG which incorporates a multiscale logits normalization distillation strategy, enhancing feature alignment, training stability and cross-view generalization.

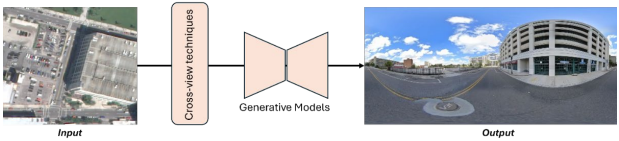

**Fig. 10.** An illustration of a typical network architecture used in cross-view synthesis methods. Most works regard it as conditional image generation task where the overhead images provide visual cues as condition, and corresponding ground-level images are generated by generative models. In addition, various cross-view techniques are also applied to bridge the domain gap between cross-view datasets.

## IV. CROSS-VIEW SYNTHESIS

Cross-view synthesis involves mapping between two distinct domains: the overhead imagery domain, which typically features low-resolution imagery capturing mainly the rooftops of buildings and the tops of trees, and the ground-level domain, which requires richer visual detail. To synthesize realistic ground-level imagery, additional information must be generated that aligns with the structural and textural patterns observed in the overhead imagery.

Existing methods generally follow the network architecture shown in **Fig. 10**, where cross-view synthesis task is formulated as the conditional image generation problem. The overhead images provide visual cues as the condition and generative models are applied to perform conditional image generation. Additionally, various cross-view techniques are proposed to bridge the domain gap between cross-view datasets.

In recent years, the rapid progress of generative models in 2D image synthesis, such as GANs [19], [103], [104], [105] and diffusion models [21], [106], [107], [108], [109], have been widely used as the backbones for cross-view generation. **Table 3** highlights representative cross-view synthesis methods, which generally shows that GANs are widely used before 2023 while diffusion models are widely used after 2023 due to their proven performance in various other vision applications.

**GANs** [19], [103], [104], [105] have achieved impressive results in image synthesis tasks. They consist of two components: a generator $G(\cdot)$ and a discriminator $D(\cdot)$. The generator $G$ creates synthetic data by taking a latent code as input, while the discriminator $D$ aims to distinguish between the generated samples and real data. During training, both networks are optimized simultaneously based on the min-max game theory [19], with the discriminator guiding the generator to produce increasingly realistic synthetic data. The objective functions of the network are defined as follows:

$$\min_G \max_D L_{GAN}(G, D) = E_{x \sim p_{data}(x)}[log D(x)] + E_{z \sim p_z(z)}\left[\log\left(1 - D(G(z))\right)\right] \quad (10)$$

where $x$ is real data sampled from data distribution $p_{data}$ and $z$ is a $d$-dimensional noise vector sampled from a Gaussian distribution $p_z$.

**Conditional GANs** [103], [104], [105], [110], [111] generate images conditioned on auxiliary information such as semantics [105], edge maps [103], or text prompts [110], [112]. In this framework, both the generator $G(\cdot)$ and discriminator $D(\cdot)$ receive a conditioning variable $c$. The generator leverages $c$ in Eq.11 to guide the image synthesis, while the discriminator evaluates pairs consisting of the conditioning variable and an image. The real input pair of $D(\cdot)$ is composed of a true image from the data distribution and its corresponding label, whereas the fake pair contains a synthesized image paired with the same label. The objective function for conditional GANs is defined as:

$$\min_G \max_D L_{cGAN}(G, D) = E_{x,c \sim p_{data}(x,c)}[log D(x, c)] + E_{x',c \sim p_{data}(x',c)}\left[\log\left(1 - D(x', c)\right)\right] \quad (11)$$

where $x' = G(z, c)$ is the generated image.

Building on the strong performance of GANs in image generation, researchers have extended these models to the task of cross-view generation [29], [113], [114], [115]. The central idea is to use GANs for conditional generation, where the conditioning inputs originate from overhead imagery processed through a prior viewpoint transformation. These conditions may include raw overhead images, ground-level semantic maps, depth information, or appearance features. In this framework, cross-view generation operates as an adversarial process: the generator learns to produce realistic ground-level images based on the overhead input, while the discriminator distinguishes between generated and real ground-level imagery. Through iterative training of both networks, GANs progressively improve the fidelity of generated ground-level imagery to closely match real-world data.

For example, Sat2Ground [29] utilizes BicycleGAN [104] to generate ground-level imagery conditioned on ground-level semantics. Unlike X-Seq [113], which incorporates uncertainty in semantics, Sat2Ground assumes these semantic maps are sufficiently accurate to be directly fed into the generator. Similarly, PanoGAN [114] applies a GAN architecture as generator for generation of both ground-level image and auxiliary semantics, where the semantics can provide additional ground-level layout information for the guidance.

**Diffusion models** are a class of generative models that synthesize data by simulating a diffusion process [106], [107], [108], [109]. The core idea is to gradually transform the original data distribution into a simple, tractable distribution, i.e., Gaussian distribution, through a forward process that adds noise over multiple steps. The model is then trained to learn the reverse process, which denoises the data step by step to recover samples that resemble those from the original data distribution.

In the forward process, noise is incrementally added to clean data until it becomes indistinguishable from pure noise. During generation, the model starts from a random noise sample $x_T \sim \mathcal{N}(0, \mathbf{I})$ and progressively denoises it into a sample $x_0 \sim p_{data}(x)$, approximating the real data distribution.

**Table 3.** Examples of cross-view generation methods are presented, categorized primarily by the type of generative model employed. The choice of scene representation and rendering technique plays a crucial role in determining how information is transferred from the overhead to the ground-level domain.

| Method | Publication | Generative Models | Scene Representations | Rendering |
|--------|-------------|-------------------|------------------------|-----------|
| Sat2Groundscape[34] | 2025 CVPR | Diffusion | Mesh | Camera projection |
| Ze et al.[116] | 2025 arXiv | Diffusion | - | - |
| Skyeyes[117] | 2025 WACV | Diffusion | Neural Rep. | Volumetric rendering |
| GPG2A[118] | 2025 WACV | Diffusion | Layout | - |
| PerLDiff[119] | 2025 arXiv | Diffusion | Layout | Camera projection |
| Crossviewdiff[120] | 2024 ICANN | Diffusion | 3D voxel | Camera projection |
| BEV2Street[121] | 2024 ICRA | Diffusion | Layout | Camera projection |
| GVG[30] | 2024 ECCV | Diffusion | Mesh | Camera projection |
| Sat2Scene[33] | 2024 CVPR | Diffusion | 3D voxel | Neural Rendering |
| BEVControl[122] | 2023 arXiv | Diffusion | Layout | Camera projection |
| PPGAN[123] | 2023 TGARS | GAN | - | - |
| Sat2Density[124] | 2023 ICCV | GAN | 3D voxel | Volumetric rendering |
| PANOGAN[114] | 2023 TMM | GAN | - | - |
| InfiniCity[125] | 2023 ICCV | GAN | 3D Voxel | Neural Rendering |
| S2SP[115] | 2022 TPAMI | GAN | 3D voxel | Panoramic Projection |
| Sat2vid[126] | 2021 ICCV | GAN | 3D voxel | Neural Rendering |
| CrossMLP[127] | 2021 BMVC | GAN | - | - |
| Toker et al[128] | 2021 CVPR | GAN | - | - |
| S2G[29] | 2020 CVPR | GAN | Height Map | Camera projection |
| LGGAN[129] | 2020 CVPR | GAN | - | - |
| H-regions[130] | 2019 CVIU | GAN | - | - |
| SelectionGAN[131] | 2019 CVPR | GAN | - | - |
| Deng et al[132] | 2018 SIGSPAT | GAN | - | - |
| X-fork[103] | 2018 CVPR | GAN | - | - |

The following section describes the forward process in more detail:

$$p(x_t|x_0) = \mathcal{N}\left(x_t; \sqrt{a_t}x_0, 1 - a_t\mathbf{I}\right) \quad (12)$$

defines the distribution of $x_t$ as a noised version of $x_0$ at diffusion step $t = 0, \dots, T$; $a_t$ is a predefined noise scheduling term.

The backward process, in contrast, iteratively denoises samples drawn from the distribution to produce the final output. It iteratively computes

$$x_{t-1} = DDIM(x_t, \epsilon_\theta(x_t, t), t) \quad (13)$$

where $\epsilon$ is a learned neural network with parameters $\theta$; $\epsilon$ predicts the noise we have added on $x_t$, which is used by a DDIM denoiser [106] to compute $x_{t-1}$.

By learning the denoising process, diffusion models can effectively model the underlying structures and patterns within data, allowing them to generate high-quality and diverse samples.

In the context of cross-view generation, GVG [30] is among the first to adopt the diffusion framework. It introduces additional learnable parameters to model geo-specific landscape styles, while encoding satellite geometry and appearance features to guide the denoising process, resulting in photorealistic ground-level view synthesis. Skyeyes [117] further advances this approach by incorporating the emerging Gaussian Splatting framework for satellite scene representation, and applying diffusion models to generate high-resolution ground-level imagery. This method demonstrates that diffusion can effectively suppress artifacts, such as needle-like white noise, commonly introduced by Gaussian Splatting when rendering scenes from wide viewing angles. Additionally, Sat2Scene [33] extends diffusion models into the 3D domain by proposing a sparse 3D diffusion framework. This model takes a satellite-derived point cloud as input and directly predicts color values for each point. As a result, the generated textures are inherently tied to the 3D structure of the scene, supporting the generation of multiple consistent ground-level imagery.

These methods typically use viewport transformation to produce guidance or conditioning inputs for the generative model to synthesize corresponding ground-level imagery. For example, S2G [29] employs BicycleGAN [104], conditioned on ground-level semantics obtained from a height map-based transformation. In contrast, GVG [30] uses a diffusion-based approach, where the conditioning comes from textured mesh

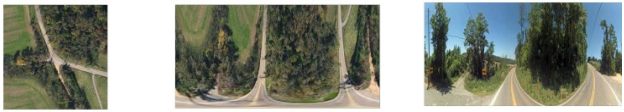

(a) Overhead view | (b) Polar-transformed view | (c) Ground truth

**Fig. 12.** Samples of polar transform.

representations derived from the overhead imagery.

**Cross-view techniques** are the critical components that differentiate various cross-view synthesis methods. They are designed to accurately transform the appearance of the scene, layout information from overhead to ground-level format to reduce the huge difference in viewing perspective and resolution in cross-view datasets. It facilitates accurate and minimally lossy information transfer despite the nearly 90-degree difference in viewpoints. Since cross-view techniques play a vital role in the performance of cross-view synthesis, we divide the existing methods based on the type of used cross-view techniques and categorize them into: End-to-end methods, Voxel-based methods, mesh-based methods, and Neural-representation-based methods.

End-to-end methods aim to bridge the cross-view domain in an end-to-end way, where the overhead images are regarded as the input of the network and the ground-level images are the direct output without intermediate results. This type of method relies on the power of generative models to learn the mapping relation between cross-view domains while they usually require large amount of training pairs to achieve reasonable performance. Recently, another type of method tends to predict the scene geometry from overhead imagery and its appearance and layout information can be transformed to ground level using projection or rendering techniques. The format of scene geometry can be 3D voxels, mesh or neural representations. We provide details for each type of method below.

### A. End-to-end methods

Early approaches to cross-view generation aimed to directly map overhead imagery to the ground-level domain without relying on intermediate representations. This could be accomplished either through mathematical warping operations that define pixel correspondences between the two views, or through end-to-end neural networks that learn a direct mapping between the domains.

**Polar transform.** The early methods [13], [64], [128] introduced the polar transform as a simple yet effective way to approximate the mapping from overhead to ground-level imagery with tolerable distortion. The core idea is that pixels along the same azimuthal direction in an overhead image roughly correspond to vertical columns in a ground-level image.

In this setup, the center of the overhead image is treated as the origin of the polar coordinate system, with the north direction (commonly available in satellite imagery) defined as the 0-degree angle. The transformed image is resized to match the dimensions of the ground-level image, denoted as $W_g \times$ $H_g$, while the original overhead image has a size of $A_a \times A_a$. The polar transform defines a mapping between the original overhead image coordinates $(x_i^s, y_i^s)$ and the transformed coordinates $(x_i^t, y_i^t)$ as shown in Eq 14,15, allowing the system to reproject overhead pixels into an approximate ground-level position.

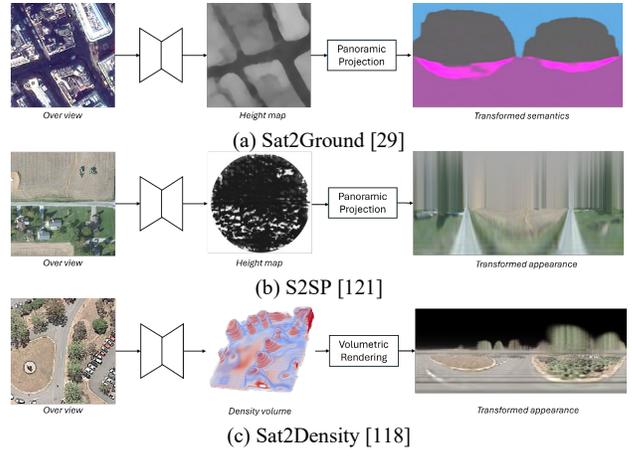

(a) Sat2Ground [29]

(b) S2SP [121]

(c) Sat2Density [118]

**Fig. 11.** Three examples of voxel-based transformation are presented. This approach assumes that the overhead imagery is captured via orthogonal projection, allowing estimated depth or density maps to be converted into 3D voxels. Ground-level images are then generated using panoramic projection or volumetric rendering techniques.

$$x_i^s = \frac{A_a}{2} + \frac{A_a}{2} \frac{y_i^t}{H_g} \sin\left(\frac{2\pi}{W_g} x_i^t\right) \quad (14)$$

$$y_i^s = \frac{A_a}{2} - \frac{A_a}{2} \frac{y_i^t}{H_g} \cos\left(\frac{2\pi}{W_g} x_i^t\right) \quad (15)$$

After applying the polar transform, objects in the transformed overhead images align roughly with their counterparts in ground-level images, as illustrated in **Fig. 11**. However, noticeable appearance distortions remain, primarily because the polar transform does not account for scene depth. For instance, elements like the sky, which are prominent in ground-level views, cannot be accurately reconstructed through this method alone.

Some studies [63], [113], [114], [127], [133] treat the conversion from overhead to ground-level imagery as an image-to-image translation task using end-to-end neural networks to learn the mapping. The approach typically follows a conditional GAN framework [103], where the generator predicts the transformed ground-level view, and the discriminator distinguishes between generated and real ground-level images. When random noise is introduced, the transformation becomes stochastic, resulting in varied outputs due to the noise input. For example, [133] is among the first to apply conditional GANs to this task. It encodes the overhead image into a feature embedding, concatenates it with a noise vector, and feeds the combined input to the generator. Similarly, X-Fork [113] employs a comparable architecture, but its generator produces both the ground-level appearance and semantic layouts, providing richer structural guidance.

PanoGAN [114] uses an encoder-decoder network to simultaneously generate the ground-level image and its corresponding segmentation map. These outputs are then evaluated by both an image discriminator and a segmentation discriminator, enabling more effective supervision signals to propagate throughout the encoder-decoder model.

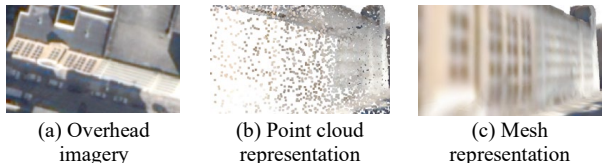

(a) Overhead imagery    (b) Point cloud representation    (c) Mesh representation

**Fig. 13.** Illustration of cross-view synthesis using point cloud and mesh representations. As shown, the mesh representation captures denser geometry and retains more detailed appearance information from the overhead imagery compared to the point cloud.

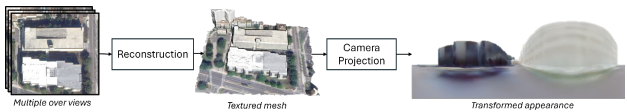

**Fig. 14.** Illustration of mesh-based transform [30], [34].

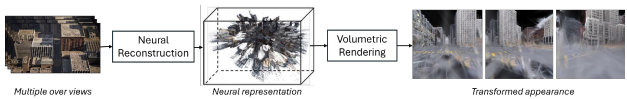

**Fig. 15.** Illustration of neural-representation-based transform.

### B. Voxel-based methods

When the viewport difference is large, an accurate scene representation is essential to minimize distortion and prevent information loss during the view transformation [134], [135]. This involves either explicitly or implicitly estimating the scene geometry, along with other details such as appearances and semantics. The ground-level views are then synthesized through rendering techniques based on this representation.

The most straightforward approach to scene modeling is to predict a height map from overhead imagery [29], [115], [124], [126]. Given an overhead image, its depth map can be estimated using convolutional neural networks, such as U-Net architectures, which are well-established in single-image depth estimation [136]. Assuming the overhead imagery is captured with an approximately **orthogonal projection**—an assumption that generally holds for satellite images taken at altitudes of 600–700 km [137]—the depth map can be converted into a height map. Next, the height map is discretized into an $n \times n \times n$ voxel grid $\mathcal{G}$ by determining whether each voxel lies above or below the corresponding height value. This grid is centered on the street-view location and incorporates the height information accordingly.

**Panoramic projection** is used to render ground-level images from the estimated occupancy grid [29], [115]. Starting from the central voxel of the grid $\mathcal{G}$, panoramic rays are cast at varying viewing angles $(\theta, \varphi)$. A ground-level panorama of size $k \times 2k$ is created by uniformly sampling $2k$ longitude angles $\theta \in [0, 2\pi]$ and $k$ latitude angles $\varphi \in [0, \pi]$, resulting in $k \times 2k$ rays penetrating the 3D voxel grid. For each ray, the depth value corresponds to the distance from the ray origin to the first occupied voxel encountered.

For example, Sat2Ground [29] estimated the depth map from the satellite imagery to initialize the scene geometry and then use the panoramic projection to transform the semantics information from overhead to ground level as the conditions. This ground-level semantics are robust and accurate to provide ground-level layout information, as illustrated in **Fig. 12-(a)**. S2SG [115] employs a similar approach to use height map as the scene geometry representation and it further transform the overhead appearance to the ground level as condition. Compared to semantics, ground-level appearance provide more details around ground surface and building facades to better guide the generative models to perform cross-view synthesis.

**Volumetric rendering** is employed to generate ground-level views when the scene is represented as a density field [23], [124], [138], as illustrated in **Fig. 12-(c)**. Unlike the occupancy grid $\mathcal{G}$, where each voxel is binary (0 or 1), the density field $\mathcal{D} \in \mathbb{R}^{H \times W \times N}$ assigns a continuous value to each voxel. The depth $d$ and opacity $\mathcal{O}$ along each queried ray can then be computed by:

$$d = \sum_{i=1}^{S} T_i a_i d_i, \quad \mathcal{O} = \sum_{i=1}^{S} T_i a_i \qquad (16)$$

where $d_i$ represents the distance from the camera to the sampled point, denotes the accumulated transmittance along the ray up to $t_i$, and $a_i$ is the alpha value used in alpha compositing, defined as:

$$a_i = 1 - \exp(-\sigma(x_i)\delta_i), \quad T_i = \prod_{j=1}^{i-1}(1 - a_j) \qquad (17)$$

For example, Sat2Density [124] applies the density field to represent the scene geometry and it employs a copy-paste strategy for image rendering, where colors are computed by sampling from the satellite image along each ray using bilinear interpolation, as shown below:

$$c_{map} = \sum_{i}^{S} T_i a_i c_i \qquad (18)$$

where $c_i = c(x_i, y_i, z_i) = I_{sat}(\frac{x_i}{S_x} + \frac{H}{2}, \frac{y_i}{S_y} + \frac{W}{2})$, with $S_x$ and $S_y$ representing the scaling factors between the satellite image's pixel coordinates and the grid coordinates in $\mathcal{D}$.

Voxel-based transformation is particularly well-suited for satellite imagery, given their approximate orthogonal projection. However, a key limitation is that height maps are a 2.5D representation, which cannot effectively capture building façade information.

### C. Mesh-based methods

Textured mesh is a full 3D scene representation that can preserve the original overhead imagery texture with minimal information loss [139], [140], [141], as shown in **Fig. 13**. Compared to voxel representation, mesh can not only represent the scene geometry but also store the building façade appearance.

**Mesh reconstruction** typically assumes multiple overhead images as input, as illustrated in **Fig. 14**. It leverages accurate camera projection models—such as Rational Polynomial Camera (RPC) models for satellite imagery [142] and perspective projection for aerial or drone imagery, to perform traditional 3D reconstruction pipelines [143], [144], [145], [146]. This process involves structure-from-motion [147], [148], dense matching [149], [150], and mesh generation [151], [152]. After reconstruction, the original overhead textures are projected onto the mesh using texture mapping techniques to create a textured mesh model [139], [153].

However, GVG [30] observed that textures near building façades often suffer from severe distortion when using the standard pipeline, primarily due to inaccuracies in the estimated mesh geometry. To address this, it introduced a texture refinement module that adjusts the mesh to produce flatter and smoother façade surfaces. Similarly, Sat2GroundScape[34] also applies the textured mesh as the scene geometry. It proposed a temporal-satellite denoising module to enable cross-view sequence synthesis from the ground-level appearance conditions.

Mesh-based transformation relies on multiple overhead images to construct a textured mesh. This approach effectively preserves both the geometric structure and appearance details of the original imagery, including subtle features like building façades, making it especially valuable for urban environments.

### D. Neural-representation-based methods

Neural scene representation methods have recently attracted significant attention in the 2D/3D generation community due to their strong ability to complete missing information and synthesize novel views [154], [155]. These methods model a scene as a learnable function $f(x)$, which encodes properties such as density, appearance, and opacity at any arbitrary 3D position within the scene. To generate ground-level images from such representations, volume rendering techniques are typically employed.

When applied to outdoor scenes, 3D Gaussian Splatting (3DGS) has demonstrated superior performance compared to traditional Neural Radiance Fields (NeRFs) [156], [157], [158], [159].

As illustrated in **Fig. 15**, when multiple overhead images are available, the neural scene representation can be optimized by minimizing the difference between the synthesized views and the corresponding input images from the same perspective. For instance, Skyeyes [117] employs 3D Gaussians to model the scene from multiple overhead views. The scene is represented as a set of differentiable 3D Gaussians, which can be efficiently rendered using tile-based rasterization. Each Gaussian is parameterized by a center point $\mu_g$ and a covariance matrix $\sum_g$:

$$G(x) = e^{-\frac{1}{2}(x-\mu_g)^T \Sigma_g^{-1}(x-\mu_g)} \quad (19)$$

During rendering, the color and opacity of each Gaussian are computed using Eq 6. The final pixel color $C$ is then obtained by blending the contributions of all 2D-projected Gaussians that influence the pixel:

$$C = \sum_{i \in N} c_i a_i \prod_{j=1}^{i-1}(1 - a_j) \quad (20)$$

where $c_i$ and $a_i$ are the view dependent on the color and opacity of the Gaussian.

## V. APPLICATION, LIMITATIONS AND FUTURE DIRECTIONS

### A. Applications

Cross-view localization and synthesis enable a wide range of applications, including improved navigation in autonomous driving, reliable positioning for robotics and drone in GPS-denied environments, and enhanced spatial awareness in augmented reality.

**Automative applications.** By integrating ground-level imagery with overhead imagery, this technology provides an advanced and efficient approach to vehicle geolocation. A notable application lies in advancing traditional vehicle navigation systems [160] . For example, when a vehicle travels through dense urban environments or winding rural roads, its navigation system can leverage this technology to align real-time camera feeds with existing satellite imagery. This alignment enables more accurate position estimation, leading to improved route planning and guidance. Furthermore, it largely accelerates the development of autonomous driving techniques. Since self-driving cars rely heavily on precise environmental data for safe and efficient navigation, integrating cross-view localization into their systems can enhance situational awareness and decision-making. By addressing challenges such as dense urban layouts and fluctuating weather or lighting conditions, this technology has the potential to significantly advance autonomous navigation [161].

**Robotics applications.** Cross-view localization demonstrates strong potential across a wide range of robotic applications, particularly in environments where GPS signals are weak or unavailable. Such scenarios include dense urban areas with tall buildings forming "urban canyons" or indoor spaces where GPS cannot penetrate. In such scenarios, models like Wide-Area Geolocalization (WAG) and Restricted FOV Wide-Area Geolocalization (ReWAG) [162] offer robust solutions for precise, real-time localization. For instance, autonomous ground robots operating in urban or indoor environments can benefit from ReWAG [162], which is tailored for robots equipped with standard (non-panoramic) cameras. By leveraging pose-aware embeddings and particle pose integration, ReWAG achieves accurate geolocation by aligning ground-level images captured by the robot with corresponding satellite or aerial imagery—enabling reliable navigation even in GPS-denied conditions.

**Drone Applications.** Traditionally, UAVs depend on GPS for navigation; however, in GPS-denied environments such as dense urban areas or large indoor spaces, localization and image synthesis become crucial. In such scenarios, drone-captured ground-level images can be matched with satellite or aerial imagery to estimate the UAV's position. A key challenge for UAVs operating in these complex settings is maintaining accurate localization despite environmental disturbances such as shadows, signal path loss, and airframe interference. Ni et al. [163] highlighted that large-scale fading (LSF), path loss (PL), and shadow fading (SF) significantly affect UAV-to-ground (U2G) communication, particularly in densely built environments. Additionally, airframe shadowing (AS) can induce power fluctuations of up to 30 dB, further complicating localization. To achieve reliable UAV positioning under such conditions, robust geolocation algorithms must effectively account for these factors.

**Augmented Reality (AR) applications**. AR has rapidly advanced as a powerful technology, enabling digital content to be seamlessly overlaid onto the physical world. A key factor in delivering truly immersive AR experiences is photorealistic view synthesis. Incorporating cross-view synthesis into AR applications holds transformative potential, enriching both spatial understanding and user engagement. A notable example is AR-based navigation [164], where cross-view synthesis can significantly enhance usability. In complex urban settings, AR can overlay intuitive, context-aware directions directly to the real world, creating a more immersive and user-friendly experience. Achieving this requires photorealistic view synthesis, where ground-level perspectives are accurately generated from aerial or satellite imagery, thereby strengthening both immersion and the reliability of navigational cues.

### B. Limitation and Future Directions

Significant progress in the quality of cross-view localization and synthesis has been driven by advancements in feature extraction backbones, generative models and cross-view techniques. Recently, these tasks have attracted considerable attention, largely due to the success of large-scale models in vision foundation models. However, numerous challenges remain before the localization accuracy and generated ground-level imagery can meet the stringent standards required for applications such as autonomous driving, video games, and immersive VR/AR content. This section explores some of the key open challenges and promising directions for future research in the field.

We first discuss the limitations and future directions in the field of cross-view localization.

**Vision foundation model.** The integration of advanced deep learning methods—particularly transformer-based architectures like BERT (Bidirectional Encoder Representations from Transformers) [165], [166], [167] and newer models such as BEiT (BERT for Image Transformations) [168], [169], [170]—holds significant promise for the field of cross-view localization. Originally developed for Natural Language Processing (NLP), transformers have demonstrated strong capabilities in handling sequential data, and this strength is increasingly being applied to image-based tasks. Transformers excel at capturing long-range dependencies and contextual relationships, regardless of the spatial or sequential distance between elements. This makes them particularly well suited for tasks which require robust contextual understanding. Consequently, transformer-based models offer a more powerful and flexible approach to addressing the inherent challenges of this task. The field is already witnessing a shift toward these architectures, as seen with models like TransGeo [54] and DSTG[44]. However, this area continues to evolve rapidly, and further advancements are expected.

**Computation cost.** The recent transformer-based models hold great promise for localization task, but their application also presents limitation is their high computational demand, which can hinder widespread adoption. Additionally, localization tasks involve handling multi-modal data—such as satellite and ground-level images—and operating at large geographical scales with high-resolution imagery, further complicating model design and deployment. To mitigate the computational burden, it is valuable to revisit classical feature- and geometry-based methods. By combining traditional feature-based strategies with modern deep learning techniques, a hybrid approach could significantly enhance accuracy in cross-view localization. Furthermore, advanced strategies like domain adaptation and multi-task learning may help address the challenges posed by the multi-modal nature of cross-view localization dataset. Domain adaptation can facilitate better transfer of learned features between satellite-view and ground-level images. Meanwhile, multi-task learning enables models to extract more generalized and robust features by training on multiple related tasks simultaneously.

Different from localization, cross-view synthesis has their unique challenge and directions as described below.

**Evaluation.** Accurately quantifying the quality of generated ground-level imagery remains an important yet underexplored issue. Common metrics like PSNR, SSIM, and FID rely on ground truth data but often fail to capture the full content quality of generated images. For example, differences in lighting conditions (such as shadows) or dynamic objects (cars, pedestrians, trees) between generated views and ground truth can significantly affect low-level metrics like PSNR, even though such differences may be less critical compared to the correct rendering of static scene elements like roads and buildings. Metrics that better simulate human perception, such as FID and DreamSIM, can assess photorealism but may not accurately reflect structural or layout correctness—an essential factor for many industrial applications. There is a need for improved metrics that objectively evaluate photorealistic quality and structural accuracy while minimizing sensitivity to irrelevant dynamic objects.

**Dataset.** Unlike language or indoor image data, which are relatively easy to capture, assembling cross-view image pairs requires coordinated collection of overhead imagery from drones, planes, or satellites alongside corresponding ground-level images. The challenges here are multifold. First, acquiring overhead imagery is costly, with few open-source satellite datasets available and commercial satellite data often prohibitively expensive. Second, aligning cross-view pairs is difficult: while GPS data provides rough alignment, it lacks precise orientation information and suffers from reduced accuracy in urban environments. Third, temporal differences between the satellite and ground-level images pose challenges; for example, datasets like DFC 2019 [39] may have months or years between captures, and Google Street View data can have similar temporal gaps. These time differences can lead to substantial scene changes, including building alterations, vegetation, vehicles, pedestrians, and lighting conditions. Consequently, a large-scale, well-aligned, and temporally synchronized cross-view dataset remains highly desirable.

**Cross-view techniques.** Converting overhead imagery to ground-level imagery is a fundamental step in cross-view generation. Various methods are discussed in Sec 3. End-to-end approaches are simple and direct but often require large-scale training data and can suffer from unstable performance. Geometry-based methods offer efficient scene representation but may struggle with inaccurate geometry estimation or require additional inputs like multiple overhead images or camera poses. For instance, Gaussian splatting techniques represent scenes from overhead images but frequently produce artifacts—such as white spikes around ground regions—that hinder providing reliable appearance and geometry information for generative models. Developing robust geometry estimation methods that preserve appearance fidelity throughout the viewport transform remains an open challenge.

**Consistent View Generation.** While generating individual ground-level views has advanced rapidly, producing multiple consistent ground-level views from overhead imagery is still nascent. Overhead images provide low-resolution appearance and weak layout cues, forcing ground-level view generation to introduce significant randomness, which complicates maintaining consistency across neighboring views. Moreover, datasets containing consistent cross-view sequences are scarce, with Sat2GroundScape being a notable exception. The ability to generate consistent sequences of ground-level views would greatly benefit real-world applications, including simulation platforms for autonomous driving and the gaming industry.

## VI. Conclusion

This paper provides a comprehensive survey of the two primary tasks in cross-view visual understanding, cross-view localization and cross-view synthesis. It reviews recent methodologies, commonly used benchmark datasets, evaluation metrics, as well as the key challenges and limitations in the field. While these two tasks share similar challenges inherent to cross-view datasets, each also faces distinct difficulties in real-world applications. We further present a systematic discussion of the core techniques employed in both tasks, including general frameworks, feature extraction backbones, and scene representation methods. The literature indicates that CNN-based approaches remain dominant in cross-view localization, although large pre-trained vision foundation models have demonstrated significant potential. In cross-view synthesis, diffusion models and their variants are currently the leading techniques; however, existing methods often struggle to generate small or fine-grained objects, emphasizing the need for more advanced approaches capable of producing realistic and detailed street-view imagery. Moreover, we provide an in-depth overview of the datasets used in cross-view research, highlighting that dataset scale and quality remain major bottlenecks in advancing current approaches. In the future, a major limitation is the reliance on a narrow set of datasets, underscoring the demand for more diverse and publicly accessible data resources to accelerate progress. Furthermore, the development of specialized evaluation metrics is necessary, as current studies largely depend on general-purpose image quality measures. Collectively, these challenges highlight that cross-view localization and synthesis remains a dynamic and rapidly evolving research area.